\documentclass[letterpaper, 10 pt, conference]{ieeeconf}  

\IEEEoverridecommandlockouts                              

\title{\LARGE \bf
Scaling Rough Terrain Locomotion with Automatic Curriculum Reinforcement Learning
}

\author{Ziming Li$^{1}$, Chenhao Li$^{1}$$^{2}$, Marco Hutter$^{1}$
\thanks{$^{1}$Robotic Systems Lab, ETH Zurich, Switzerland 
        {\tt\small \{zimili, chenhli, mahutter\}@ethz.ch}}%
\thanks{$^{2}$ETH AI Center, ETH Zurich, Switzerland}      
}

\usepackage{amsmath,amssymb,amsfonts}
\usepackage{hyperref}
\usepackage{graphicx}
\usepackage{booktabs}
\DeclareMathAlphabet{\mathbbold}{U}{bbold}{m}{n}
\usepackage[caption=false]{subfig}
\usepackage{graphicx}
\begin{document}

\maketitle
\thispagestyle{empty}
\pagestyle{empty}
\begin{abstract}
Curriculum learning has demonstrated substantial effectiveness in robot learning. However, it still faces limitations when scaling to complex, wide-ranging task spaces. Such task spaces often lack a well-defined difficulty structure, making the difficulty ordering required by previous methods challenging to define. We propose a Learning Progress-based Automatic Curriculum Reinforcement Learning (LP-ACRL) framework, which estimates the agent's learning progress online and adaptively adjusts the task-sampling distribution, thereby enabling automatic curriculum generation without prior knowledge of the difficulty distribution over the task space. Policies trained with LP-ACRL enable the ANYmal D quadruped to achieve and maintain stable, high-speed locomotion at 2.5 m/s linear velocity and 3.0 rad/s angular velocity across diverse terrains, including stairs, slopes, gravel, and low-friction flat surfaces--whereas previous methods have generally been limited to high speeds on flat terrain or low speeds on complex terrain. Experimental results demonstrate that LP-ACRL exhibits strong scalability and real-world applicability, providing a robust baseline for future research on curriculum generation in complex, wide-ranging robotic learning task spaces.
\end{abstract}

\section{Introduction}
Enabling legged robots to traverse complex, unstructured environments with agility and versatility remains a grand challenge in robotics. 
Curriculum Reinforcement Learning (CRL) has emerged as a key methodology for addressing this challenge by structuring the training process as a progression from simpler to more difficult tasks, thereby enabling the progressive acquisition of high-demand capabilities. 
In legged robotics, CRL has enabled policies that can traverse diverse and challenging terrains at moderate speeds and achieve high-speed locomotion on near-flat terrain.
Mainstream CRL applications~\cite{hoellerANYmalParkourLearning2023,liLearningAgilityAdaptive2023,margolisRapidLocomotionReinforcement2022,rudinLearningWalkMinutes2022,xieALLSTEPSCurriculumdrivenLearning2020}  predominantly rely on manually designed curricula, in which the agent is exposed to a predefined sequence of tasks ordered along a single difficulty axis, with advancement gated by hand-tuned performance thresholds.

However, in real-world scenarios, robots are expected to perform a diverse set of tasks that do not adhere to a well-defined difficulty ordering. 
For instance, comparing the difficulty of slow walking on stairs and fast running on gravel is inherently ambiguous. 
These tasks collectively form an unstructured task space, in which the absence of a clearly ordered sequence of tasks makes a manually designed curriculum infeasible.
In this context, Automatic Curriculum Learning (ACL)~\cite{gravesAutomatedCurriculumLearning2017} leverages learning dynamics to automatically select tasks suited to the agent’s current competence, removing the need for prior knowledge of task space difficulty and offering strong potential for scalable robot learning.

In this work, we introduce a Learning Progress-based Automatic Curriculum Reinforcement Learning (LP-ACRL) framework. 
LP-ACRL continuously reallocates sampling toward tasks estimated to provide higher learning progress~\cite{matiisenTeacherStudentCurriculumLearning2017,portelasTeacherAlgorithmsCurriculum2019}, as inferred from the episodic reward, thereby scaling robot learning to complex unstructured task spaces and enabling robots to perform a broader range of tasks in the real world.

\begin{figure}[t]
\centering
\includegraphics[width=\linewidth]{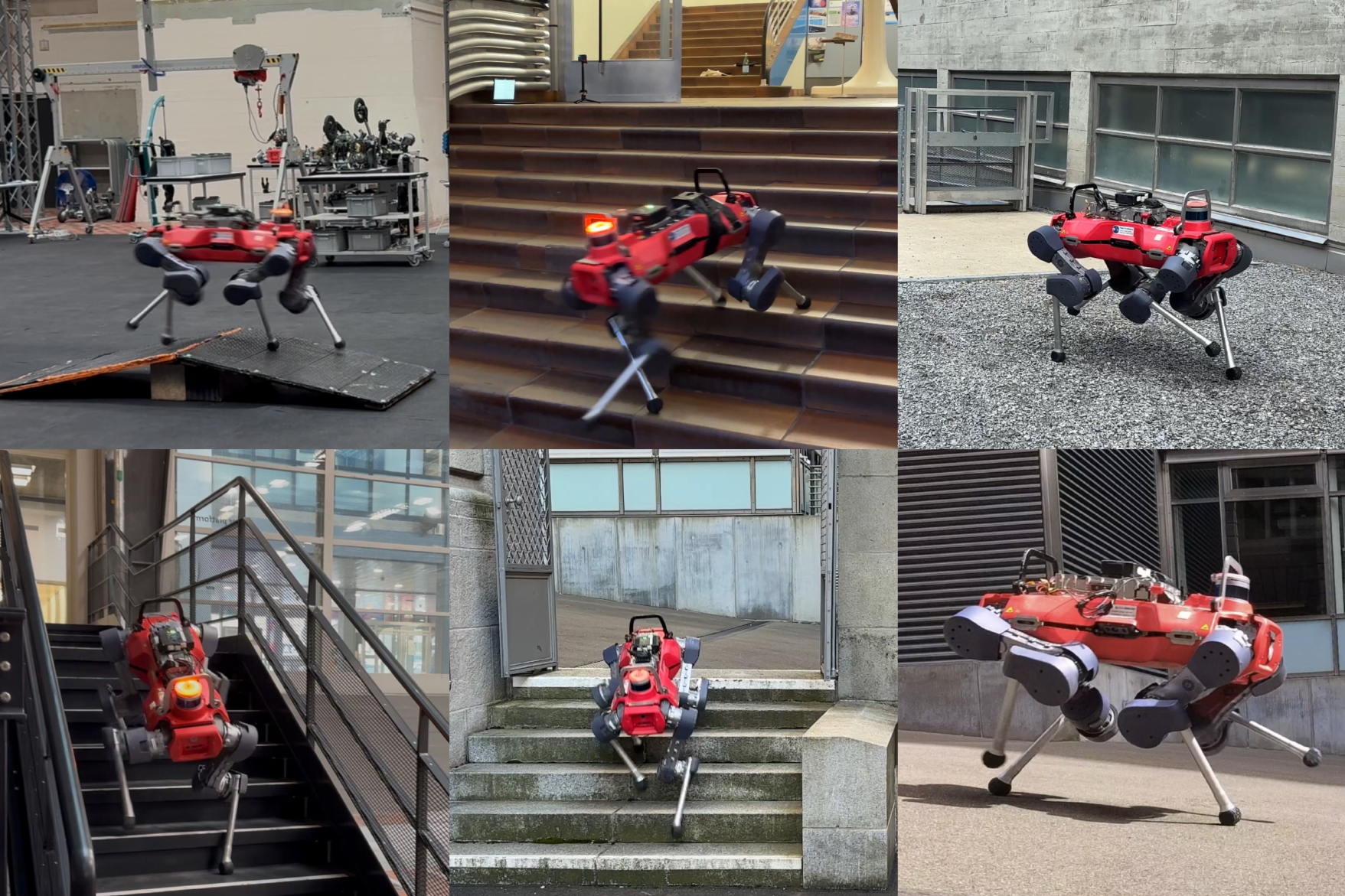}
\caption{Real World Deployment. Policy trained with \textbf{LP-ACRL} enables the ANYmal D robot to traverse diverse rough terrains, 
achieving speeds up to 2.5\,m/s and 3.0\,rad/s.  Supplementary videos and implementation details are available on our project webpage \url{https://sites.google.com/view/lp-acrl}.}
\label{fig:head}
\end{figure}

Across all our experiments, LP-ACRL consistently outperforms baselines, including multiple handcrafted and automatic CRL methods.
In the scaled locomotion task, our method rapidly achieves high success and strong performance, while existing methods fail to attain comparable results even with extended training.

Finally, our policy successfully transfers to the physical ANYmal-D platform via Teacher-Student Distillation, 
exhibiting robust locomotion capabilities. 
A single policy covers a wide range of velocity tracking tasks across diverse terrains, achieving linear velocities of $3.0~\mathrm{m/s}$ on flat terrain and $2.5~\mathrm{m/s}$ on challenging terrains such as stairs, slopes, and gravel, as well as angular velocities up to $3.0~\mathrm{rad/s}$ across all listed terrains.
To the best of our knowledge, our method achieves the highest rough terrain locomotion speed on this hardware.

Our primary contributions include:
\begin{itemize}
    \item We propose LP-ACRL, an automatic curriculum reinforcement learning framework leveraging learning progress metrics derived from episodic rewards to dynamically adjust task sampling, removing the necessity of manually constructed curricula for scalable robot learning.

    \item We validate LP-ACRL across three locomotion task spaces, including multi-level velocity tracking on flat terrain, diverse rough terrain traversal, and a large-scale locomotion task space integrating multiple linear and angular velocity levels, terrain types, and terrain levels.

    \item We conduct a comprehensive real-world evaluation on the ANYmal D hardware, demonstrating high-speed rough terrain locomotion capability ever achieved on this platform.
\end{itemize}

\section{Related Work}
\subsection{Curriculum Reinforcement Learning in Legged Robotics}
Curriculum Reinforcement Learning in legged robotics is most commonly utilized to enhance locomotion capabilities across diverse terrains.
A mainstream research paradigm employs manually designed terrain curricula during training.
For example, Rudin et al.~\cite{rudinLearningWalkMinutes2022} distribute parallel robots across various predefined terrain types. 
The curriculum then progresses by moving each robot, controlled by the same policy, independently to a more or less difficult, pre‑generated terrain within the same type, once it meets a performance threshold.
A similar method is applied for various terrain traversal in~\cite{hoellerANYmalParkourLearning2023}, with dedicated curricula designed for training expert policies mastering each terrain type, which imposes significant engineering effort and fails to scale to arbitrary terrain setups.
For velocity tracking, Ji et al.~\cite{jiConcurrentTrainingControl2022} use a simple approach of increasing the upper limit for sampling linear‑velocity commands as the number of training steps increase. 
However, manually designed methods are typically limited to a single difficulty dimension, such as the terrain geometry scale or the linear velocity command range, whose increase intuitively reflects the sequence of learning difficulty.
To address the coupling of non‑single dimensions, Margolis et al.~\cite{margolisRapidLocomotionReinforcement2022} propose a grid-expansion method based on reward thresholds to simultaneously extend the target ranges of both linear and angular velocities. 
Meanwhile, Li et al.~\cite{liLearningAgilityAdaptive2023} forcibly discretize multiple difficulty axes, such as velocity, domain randomization parameters, and behavioral penalties in the reward function, into a single ten-level difficulty axis to control curriculum progression.

However, these methods share a common limitation: their curriculum design heavily relies on prior knowledge of task difficulty and expected performance. 
When a robot needs to adapt simultaneously to different terrains and a wide range of velocity commands, the task space is no longer single‑dimensional and can scarcely be compressed into a single axis. 
The coupling between these dimensions forms a highly complex multi-axis task space, making it nearly impossible to design an effective manual curriculum.

\subsection{Automatic Curriculum Learning}
Automatic Curriculum Learning offers a principled alternative to address this challenge. 
Mainstream ACL methods can be broadly categorized into three types based on the mechanism used to evaluate task value. 
The first, widely used in game‑playing tasks, is based on adversarial methods such as AlphaGo Zero~\cite{silverMasteringGameGo2017} and AlphaStar~\cite{vinyalsGrandmasterLevelStarCraft2019}, which optimize curriculum sampling through a self‑play-like process. 
The second category is based on the learning progress (LP). 
Methods such as Mysore et al.~\cite{mysoreRewardguidedCurriculumRobust}, Portelas et al.~\cite{portelasTeacherAlgorithmsCurriculum2019}, and Matiisen et al.~\cite{matiisenTeacherStudentCurriculumLearning2017} identify the fastest learning regions by monitoring changes in rewards or success rates over time; 
absolute learning progress is also employed to mitigate catastrophic forgetting~\cite{portelasTeacherAlgorithmsCurriculum2019}.
The third category is based on “surprise” or “potential”.
Approaches such as those of Schaul et al.~\cite{schaulPrioritizedExperienceReplay2016} and Jiang et al.\cite{jiangPrioritizedLevelReplay2021} use the temporal‑difference (TD) error to measure cognitive uncertainty about task difficulty, thereby guiding the agent to prioritize exploration in highly uncertain regions. 
All three categories demonstrated strong learning efficiency and generalization capabilities on multiple reinforcement‑learning benchmarks.

In the field of legged robotics, one of the earliest explorations of ACL dates back to the work of Lee et al.~\cite{leeLearningQuadrupedalLocomotion2020}, which uses particle filters combined with online performance information to automatically generate multi-axis terrain parameters. 
Although this method still relies on manually set evaluation thresholds to decide sampling in a given parameter region, it validates the feasibility of achieving adaptive curriculum adjustment on real robots based on interaction data in a complex multi-axis task space.
Recent work by Li et al.~\cite{liFLDFourierLatent2024c} further verified the effectiveness of ACL in legged robot locomotion. 
This method compresses target skill trajectories into a low-dimensional latent space composed of frequency parameters and combines it with the ALP-GMM algorithm~\cite{portelasTeacherAlgorithmsCurriculum2019} based on absolute learning progress to adaptively sample high-value targets in this latent space. 
This approach not only significantly expands the robot locomotion skill repertoire but also effectively avoids unlearnable regions in policy training, thereby enhancing exploration efficiency while preserving already mastered skills.
While existing ACL approaches have demonstrated promising results, extending them to efficiently scale to complex, unstructured locomotion task spaces remains a challenging research problem.

\section{Approach}
\subsection{Problem Formulation}
\subsubsection{Definition of Curriculum}
The task space is denoted by $\mathcal{T}$. 
A curriculum $\mathcal{C}$ is defined as a sequence of task-sampling distributions:
\begin{align}
    \mathcal{C} = \left(c_0, c_1, \ldots, c_j, \ldots\right), \quad c_j \in \Delta(\mathcal{T}),
\end{align}
where each $c_j$ represents the task-sampling distribution over $\mathcal{T}$ employed to select training tasks at curriculum stage $j$.
\subsubsection{Markov Decision Process}
A Markov Decision Process (MDP) is formalized as
\begin{align}
\mathcal{M}=(\mathcal{S}, \mathcal{A}, \mathcal{P}, \mathcal{R}, \gamma),
\end{align}
where $\mathcal{S}$ denotes the state space, 
$\mathcal{A}$ is the action space, 
$\mathcal{P}: \mathcal{S} \times \mathcal{A} \rightarrow \mathcal{S}$ is the state transition function, 
$\mathcal{R}: \mathcal{S} \times \mathcal{A} \rightarrow \mathbb{R}$ is the reward function, 
with the reward $r_t=\mathcal{R}\left(s_t, a_t\right)$, and $\gamma \in[0,1]$ is the discount factor. 
The agent's objective is to learn a policy $\pi_\theta: \mathcal{S} \rightarrow \mathcal{A}$ that maximizes the expected discounted return $\mathbb{E}_{\pi_\theta}\left[\sum_{t \geq 0} \gamma^t r_t\right]$.

\subsubsection{Policy Update Guided By Curriculum}
A learning agent begins with an initial policy $\pi_{\theta_0}$ parameterized by $\theta_0$ and iteratively updates its parameter over discrete steps $k=0,1,2, \ldots$.
At each interaction step within each curriculum stage $j$, the agent, following its policy $\pi_\theta(a \mid s)$, interacts with the environment under the task-sampling distribution $c_j$, 
thereby generating the trajectory 
\begin{align} \tau(\theta, c_j)=\bigl(s_0,a_0,r_0,s_1,a_1,r_1,\ldots\bigr).\end{align}

The dataset $\mathcal{D}_k=\{\tau_i\}$ aggregates multiple such trajectories.
At policy update iteration $k$, the parameter is updated by the RL algorithm $\mathcal{G}$:
\begin{align}
\theta_{k+1}=\mathcal{G}\left(\theta_k, \mathcal{D}_k\right).
\end{align}

The curriculum $\mathcal{C}$ influences learning by adjusting the task-sampling distribution $c_j$. 
This shapes the statistical properties of the trajectories $\tau$, 
which in turn determine the dataset $\mathcal{D}_k$ and ultimately guide the evolution of the policy parameters $\theta$ in parameter space.

\subsubsection{Objective of Automatic Curriculum Reinforcement Learning}

Given a task space $\mathcal{T}$ without prior structural assumptions, some tasks may be intrinsically too difficult for the agent to master, forming an unmasterable subspace. 
Consequently, for any final policy, the task space can be partitioned into a mastered subspace $\mathcal{T}_{\text{mastered}}$ and an unmastered subspace $\mathcal{T}_{\text{unmastered}}$.  

The objective of Automatic Curriculum Reinforcement Learning is to find a curriculum $\mathcal{C}$ that optimizes the final policy’s overall competence, with two sub-goals:
\begin{enumerate}
\renewcommand{\labelenumi}{(\arabic{enumi})}
\item Minimizing the size of the unmastered subspace $\mathcal{T}_{\text{unmastered}}$, thereby expanding the breadth of the agent’s capabilities.
    \item Maximizing the performance of the final policy across the mastered subspace $\mathcal{T}_{\text{mastered}}$.

\end{enumerate}

\subsection{Methodology}

We propose a Learning Progress-based Automatic Curriculum Reinforcement Learning (LP-ACRL) framework.
The central idea is to quantify the agent’s learning progress on each task and to adapt the task-sampling distributions $c_j$ to prioritize tasks with the highest potential for improvement.

\subsubsection{Discrete Task Space}
Our method operates on a discrete task space $\mathcal{T}$. 
Each task instance $\zeta \in \mathcal{T}$ is characterized by multiple task dimensions, which can be either:
\begin{enumerate}
\renewcommand{\labelenumi}{(\arabic{enumi})}
    \item \textbf{Categorical dimensions}: distinct and mutually independent options that cannot be represented or parameterized in a continuous form, such as different terrain types for robot locomotion.
    \item \textbf{Continuous dimensions}: parameters that vary over bounded ranges and can be represented or parameterized in a continuous form, such as the target velocity commands that the robot is expected to track.
\end{enumerate}

All continuous dimensions are discretized by partitioning their value ranges into non-overlapping sub-intervals.
Thus, a discrete task instance $\zeta$ is defined by specific categorical choices and specific sub-intervals for continuous parameters.

\subsubsection{Learning Progress-based Automatic Curriculum Reinforcement Learning (LP-ACRL)}
\begin{figure}[h]
    \centering
\includegraphics[width=\linewidth]{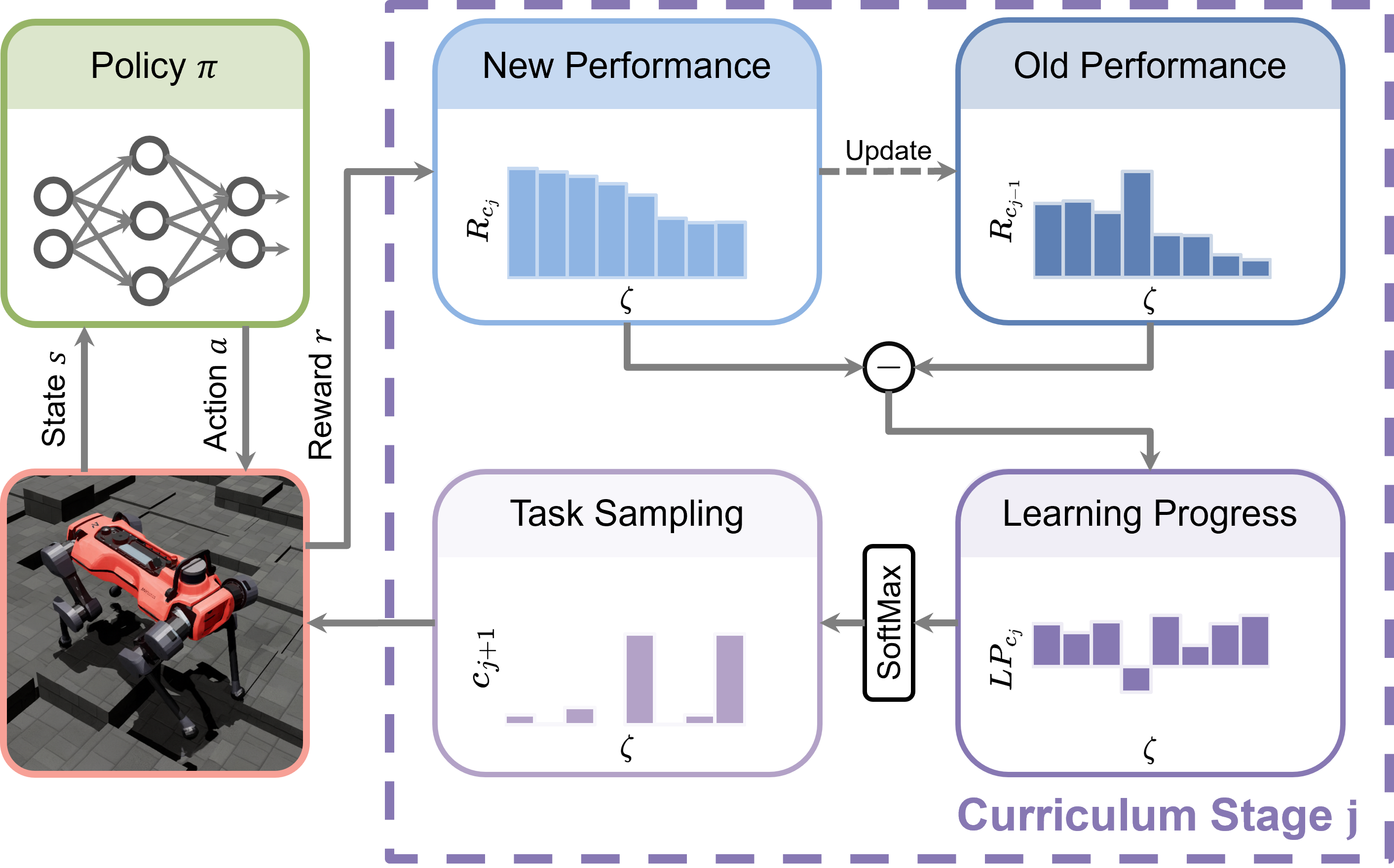}
    \caption{Learning Progress-based Automatic Curriculum Reinforcement Learning.The task-sampling distribution is updated based on learning progress, prioritizing policy learning on the most informative tasks for further improvement. }
            \label{fig:lpacrl}
\end{figure}

The core mechanism of LP-ACRL lies in its adaptive adjustment process, driven by learning progress (LP) values estimated from the agent’s performance on individual task instances. These LP values are used to update the task-sampling distribution, determining the sampling probability for each task instance. In this way, LP-ACRL prioritizes tasks of suitable difficulty that are most likely to promote effective exploration and learning by the agent.
The complete workflow of LP-ACRL is shown in Figure~\ref{fig:lpacrl}.

The agent’s performance on a trajectory $\tau = (s_0, a_0, r_0, \ldots, s_{H-1}, a_{H-1}, r_{H-1})$ of length $H$ is measured by its episodic reward $R_{\tau}$.
For a task instance $\zeta$, the expected episodic reward is estimated from trajectories:
\begin{align}
R_{c_j}(\zeta)=\mathbb{E}_{\tau \sim c_j}[R_\tau].
\end{align}
The learning progress $LP$ on $\zeta$ is defined as the change in average episodic reward between two consecutive evaluations:
\begin{align}
L P_{c_j}(\zeta)=R_{c_j}(\zeta)-R_{c_{j-1}}(\zeta).
\end{align}
For each curriculum stage, the task-sampling distribution $c$ is updated by a softmax operator over the most recent learning-progress estimates. 
$\beta$ is a temperature parameter that controls the sharpness of the distribution.
\begin{equation}
    c_{j+1}(\zeta)=\frac{e^{L P_{c_j}(\zeta) / \beta}}{\sum_{\zeta^{\prime} \in \mathcal{T}} e^{L P_{c_j}(\zeta^{\prime} ) / \beta}}.
\end{equation}

\section{Experiments}
We conduct a series of legged-locomotion experiments to evaluate the effectiveness, robustness, and scalability of the proposed LP-based framework.
We begin with linear-velocity tracking tasks on flat terrain to enable controlled comparisons.
We then consider a task space involving locomotion over diverse terrain types.
Finally, we scale up to a large, heterogeneous task set that jointly varies linear and angular velocity commands, terrain types, and terrain difficulty (parameterized by terrain geometry), 
characterizing unstructured scenarios with limited task difficulty knowledge.
The policy trained with our framework on the scaled task space is subsequently deployed on a real robot.
All simulations are performed in IsaacLab~\cite{mittal2023orbit},
and the ANYmal D\cite{hutter2016anymal} quadruped robot with 12 degree-of-freedom joints is used for both simulation and real-world deployment.
The basic locomotion training framework, including the default observation space, action space, and reward functions, is provided in Appendix~\ref{apd:locomotion framework}.

\subsection{Baselines}
We compare our \textbf{LP-ACRL} with Absolute Learning Progress (\textbf{ALP})~\cite{portelasTeacherAlgorithmsCurriculum2019}, Prioritized Level Replay (\textbf{PLR})~\cite{jiangPrioritizedLevelReplay2021}, a Simple Hand-Crafted Curriculum (\textbf{SC})~\cite{jiConcurrentTrainingControl2022}, a Low-Reward Prioritized Curriculum (\textbf{LRPC}), and Uniform Sampling (\textbf{Uniform}). 
Detailed descriptions of the baseline implementations are provided in Appendix~\ref{apd:baseline methods}.
\subsection{Episodic Percentage Tracking Error with Stability Penalty}
\label{ss:EPTE-SP}
To provide a more interpretable evaluation metric, we introduce the Episodic Percentage Tracking Error with Stability Penalty (EPTE-SP). 
This metric jointly evaluates velocity tracking accuracy and stability, explicitly penalizing early terminations due to falls. 
It is defined as
\begin{align}
\tilde{\varepsilon} = 
\frac{\varepsilon k_f +  (K - k_f)}{K},
\end{align}
where $K$ is the total episode length in steps;
$k_f$ is the first step at which the robot falls ($k_f=K$ if no fall occurs);
$\varepsilon$ is the percentage tracking error computed over the non-falling portion; 
and $\tilde{\varepsilon}$ is EPTE-SP, the percentage tracking error aggregated over the entire episode with a stability penalty.

In this formulation, the first term \(\varepsilon k_f\) accumulates the tracking error over the non-falling portion of the episode, while the second term \((K - k_f)\) assigns the worst-case error to all remaining steps after a fall. 
The sum is normalized by the episode length \(K\) to yield a percentage over the full episode duration. 
A lower \(\tilde{\varepsilon}\) indicates superior joint performance in both tracking accuracy and locomotion robustness.

\subsection{Multi-Level Linear-Velocity Tracking on Flat Terrain}
\label{ss:lin_vel}
We first validate and analyze our method and baselines in a structured task space. In this experiment, the robot aims to track the commanded linear velocity on flat terrain over a wide range.
Owing to the robot’s symmetry,
commands $\mathbf{v}_{b,x}^*$ and $-\mathbf{v}_{b,x}^*$ are treated as the same task instance \cite{mittal2024symmetryconsiderationslearningtask}.
The task space is defined over the absolute commanded linear velocity $|\mathbf{v}_{b, x}^*| \in[0,4.0]\,\text{m/s}$, discretized into eight task instances of width $0.5\,\text{m/s}$. 
The sign of $\mathbf{v}_{b, x}^*$ is sampled uniformly, while the lateral velocity command $\mathbf{v}_{b, y}^*$ and yaw-rate command $\boldsymbol{\omega}_{b, z}^*$ are fixed at zero.

\begin{figure*}[th!]
    \centering
    \includegraphics[width=\linewidth]{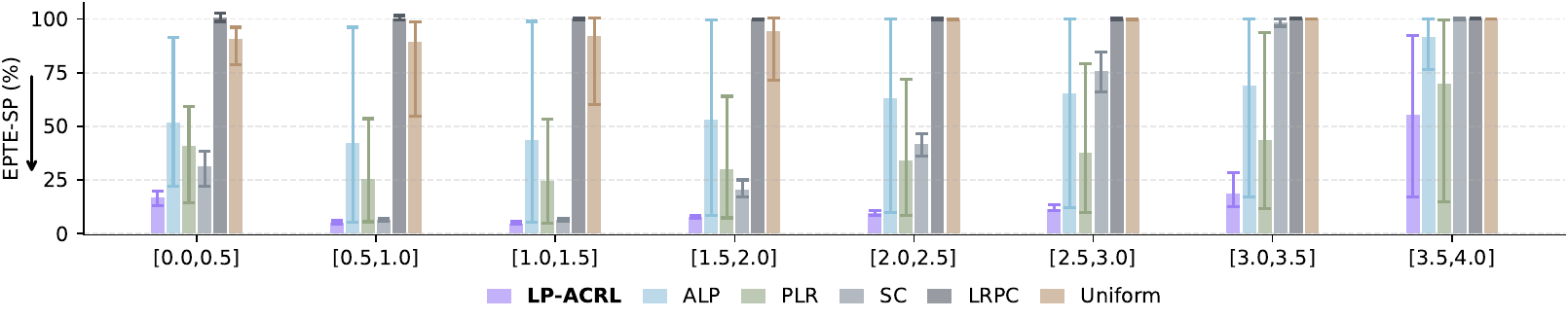}
    
    \caption{Episodic Percentage Tracking Error with Stability Penalty (EPTE-SP, $\tilde{\varepsilon}$) evaluation. The horizontal axis represents task instances, defined by the range of the absolute commanded linear velocity (in m/s). Bars indicate the mean performance averaged over multiple random seeds, with error bars representing the minimum and maximum values. The results demonstrate that the policy trained via \textbf{LP-ACRL} significantly outperforms all baseline policies,consistently maintaining the lowest EPTE-SP and narrowest variability across most intervals, while preserving its superiority in mean performance even in the most challenging task instance. }
    \label{fig:speed_error_bins}
\end{figure*}

\begin{figure}[h!]
    \centering
    \includegraphics[width=\linewidth]{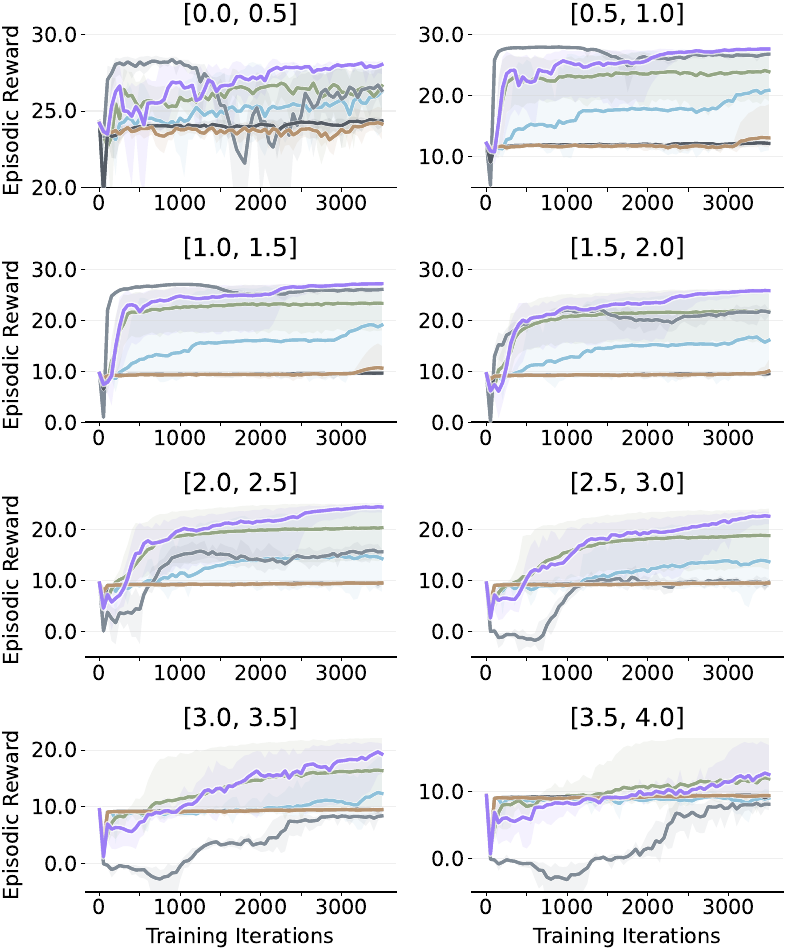}
\includegraphics[width=\linewidth]{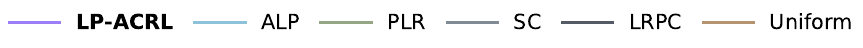}
    \caption{Evolution of Episodic Reward throughout the training process. Subplots represent distinct task instances, defined by the range of the absolute commanded linear velocity (in m/s). The horizontal axis denotes training iterations. Solid curves indicate the mean reward averaged over multiple random seeds, with shaded regions indicating the variability. The results demonstrate that the policy trained via \textbf{LP-ACRL} exhibits significantly faster convergence and superior asymptotic performance compared to baselines across the task instances.}
    \label{fig:reward_tasks}
\end{figure}

\begin{figure}[h!]
    \centering
    
\includegraphics[width=\linewidth]{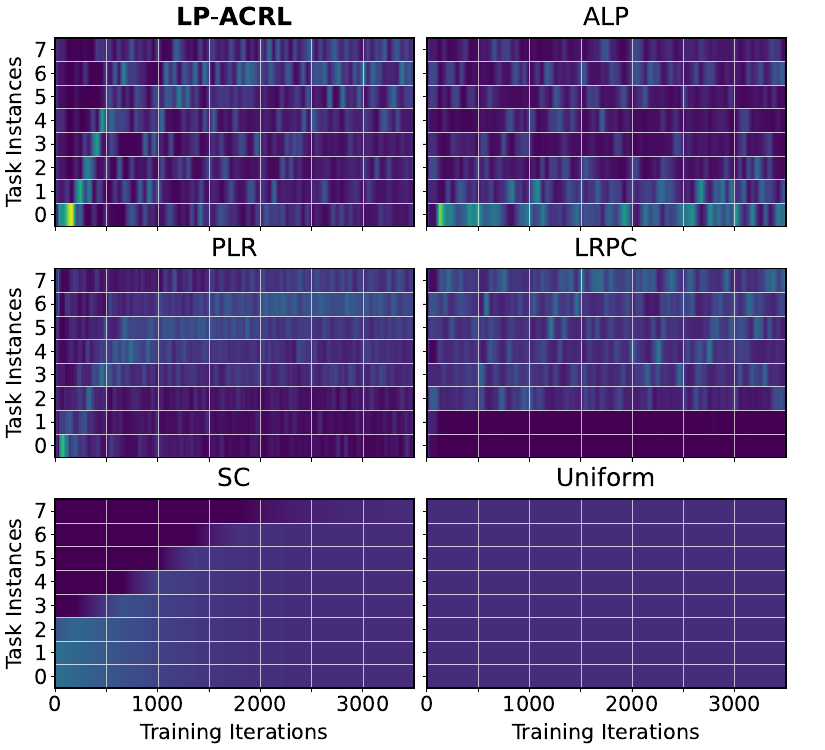}
\includegraphics[width=\linewidth]{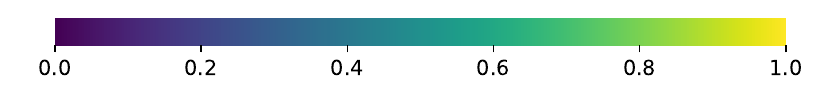}

    \caption{Task-sampling distribution under different curriculum methods. The vertical axis denotes the eight task instances (indices 0-7), corresponding to the eight velocity tracking ranges.
    The colormap indicates the sampling probability of each task instance. 
    \textbf{LP-ACRL} rapidly infers the underlying difficulty structure early in training, subsequently modulating sampling to balance high-difficulty exploration with the retention of mastered tasks.}
    \label{fig:allocation_heatmaps}
\end{figure}

We observe that our method, \textbf{LP-ACRL}, achieves the best overall performance in terms of both EPTE-SP (Fig.~\ref{fig:speed_error_bins}) and reward convergence (Fig.~\ref{fig:reward_tasks}).
An inspection of the task sampling distribution shift (Fig.~\ref{fig:allocation_heatmaps}) further demonstrates the ability of \textbf{LP-ACRL} to rapidly identify the difficulty structure of the task space at an early stage of training and to guide the agent to progress gradually from easy to difficult tasks.
At the beginning of training, it focuses on the tasks with the highest learning progress, which are typically the easiest ones.
The resulting performance improvement then transfers to slightly more difficult tasks, enabling a smooth transition of the sampling distribution toward harder tasks as the policy’s capability grows.
When learning on the hardest task instances reaches a plateau, the sampling probability naturally shifts back to medium- and low-difficulty tasks that still exhibit positive learning progress, thereby stabilizing the skills already acquired.

In contrast, both \textbf{Uniform} and \textbf{LRPC} allocate samples to task instances where the robot consistently fails, resulting in data that provide little informative learning signal and thus fails to improve performance.
\textbf{SC} achieves rapid improvement under low commanded velocities (Fig.~\ref{fig:reward_tasks}), but once higher commanded velocities are introduced, failures on high-difficulty tasks contaminate the data used for policy updates, while the sampling proportion of easy and medium tasks decreases.
This leads to severe performance degradation on those task instances.
\textbf{ALP} assigns task-sampling weights based on the absolute value of learning progress, so both improvements and regressions increase sampling probability. 
However, due to large performance fluctuations on difficult task instances early in training, \textbf{ALP} under-focuses on easy ones at first and later over-emphasizes oscillations in those easy tasks (Fig.~\ref{fig:allocation_heatmaps}), thereby failing to explore the task space effectively.
\textbf{PLR} relies on a value-prediction-error signal which provides limited task-level discrimination and is easily influenced by small numerical noise, leading to large variance in the resulting policy performance.

\subsection{Multi-Type Rough Terrain Locomotion}
We extend the evaluation of LP-ACRL on more complex rough terrain locomotion, where the task space is defined by diverse terrain types with no difficulty structure.
The terrain set consists of six types: ascending stairs, descending stairs, uphill slope, downhill slope, random rough terrain, and flat ground.
For each task instance, the commanded velocities are uniformly sampled within the ranges 
$\lvert \mathbf{v}_{b,x}^* \rvert \in [0,1]$, 
$\lvert \mathbf{v}_{b,y}^* \rvert \in [0,1]$, 
and $\lvert \boldsymbol{\omega}_{b,z}^* \rvert \in [0,1]$.

\begin{figure}[thb]
    \centering
    \includegraphics[width=\linewidth]{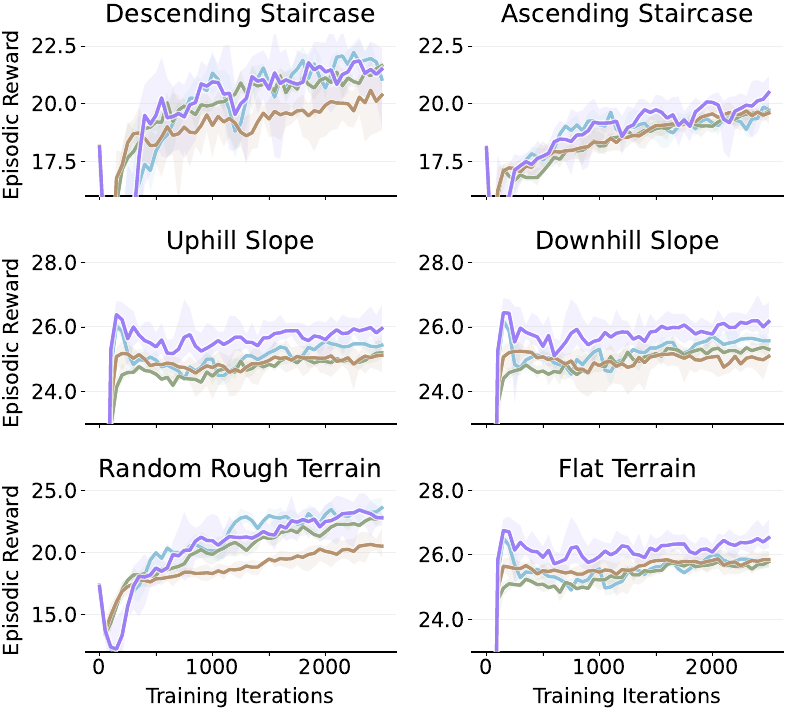}
    \includegraphics[width=\linewidth]{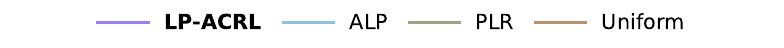}
    
    \caption{Evolution of Episodic Reward throughout the training process. Subplots represent distinct task instances, defined by different terrain type. The horizontal axis denotes training iterations. Solid curves indicate the mean reward averaged over multiple random seeds, with shaded regions indicating the variability. The results demonstrate that, even in relatively simple task spaces, \textbf{LP-ACRL} outperforms baselines in terms of both convergence rate and final performance.}
    \label{fig:b1}
\end{figure}

\begin{figure}[thb]
    \centering
    \includegraphics[width=\linewidth]{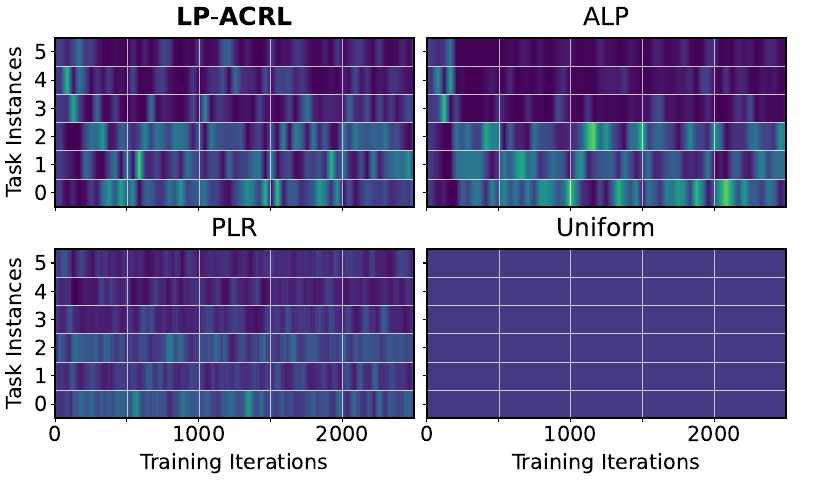}
\includegraphics[width=\linewidth]{figs/experiment/experiment_a/Task-sampling_distribution_cbar.pdf}
    \caption{Task-Sampling Distribution Heatmaps. Terrain type index (0-5) only indicates the category and does not imply difficulty:
    0: Descending Staircase,
    1: Ascending Staircase,
    2: Uphill Slope,
    3: Downhill Slope,
    4: Random Rough Terrain,
    5: Flat Terrain.
    \textbf{LP-ACRL} distinguishes itself by effectively capturing the underlying difficulty structure, while avoiding the over-fixation on specific outlier tasks.}
    \label{fig:b3}
\end{figure}

Consistent with Sec.~\ref{ss:lin_vel}, the superior performance of \textbf{LP-ACRL} is demonstrated on tasks even without clearly structured difficulty patterns (Fig.~\ref{fig:b1}).
In Fig.~\ref{fig:b3}, \textbf{LP-ACRL} effectively captures the difficulty structure during the early stages of training and maintains the flexibility to adaptively schedule across tasks in the mid-to-late stages.
In contrast, \textbf{PLR} exhibits limitations similarly indicated in the previous experiment, where the discriminability of its signals across task instances is notably weak.
Meanwhile, \textbf{ALP} remains hampered by performance fluctuations in individual tasks, thereby failing to enhance overall performance.

\subsection{Scaled Locomotion}

This experiment evaluates the scalability of our framework in a large-scale and challenging locomotion task space that combines multiple levels of linear and angular velocity tracking, diverse terrain types, and varying terrain difficulties.

\begin{figure*}[h!]
    \centering
    \includegraphics[width=\linewidth]{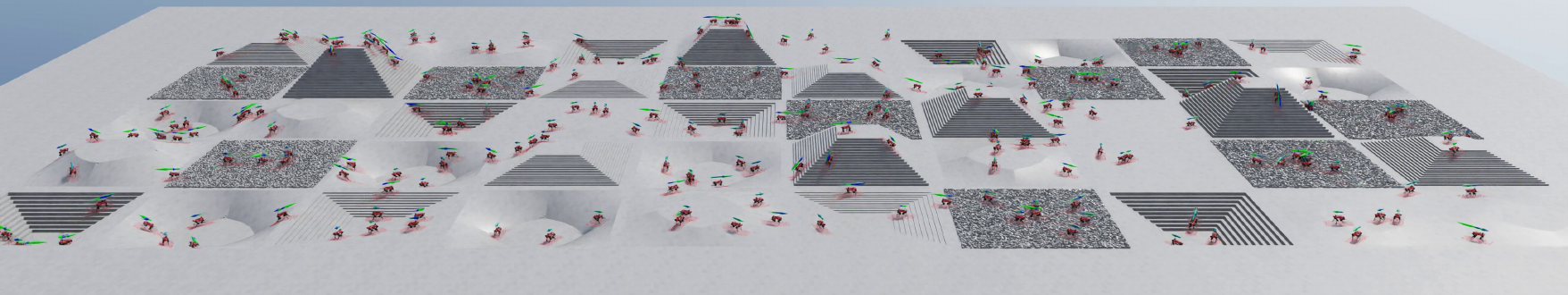}
    \caption{Illustration of the scaled locomotion task space. The task space combines diverse terrain types and geometry parameters with multiple levels of commanded linear and angular velocity. The task difficulty is unknown during policy training. The green arrow indicates the commanded base velocity, and the blue arrow indicates the robot's base velocity. The policy takes proprioception and local terrain height samples (red dots around the robot) as input. With \textbf{LP-ACRL}, the policy achieves an $80\%$ success rate in this large-scale unstructured task space within 1500 training iterations whereas the baseline methods still struggle to converge.}

\end{figure*}

The commanded linear velocity along the $x$-axis is set within $\lvert \mathbf{v}_{b,x}^* \rvert \in [0,2.5]~\text{m/s}$, discretized into five levels with an interval of $0.5~\text{m/s}$.
The commanded linear velocity along the $y$-axis is sampled from a narrower range $\lvert \mathbf{v}_{b,y}^* \rvert \in [0,0.5]~\text{m/s}$.
The commanded angular velocity around the $z$-axis is within $\lvert \boldsymbol{\omega}_{b,z}^* \rvert \in [0,3.0]~\text{rad/s}$, discretized into six levels with an interval of $0.5~\text{rad/s}$.
The terrain set includes ascending stairs, descending stairs, random rough terrain, uphill slopes, and downhill slopes. 
Each terrain type contains four difficulty levels determined by geometric parameters, as detailed in Appendix~\ref{apd:terrain parameters}.
The Cartesian product of linear and angular velocity levels, terrain types, and terrain difficulty levels defines a large-scale task space comprising 600 task instances. 
The resulting space exhibits an unstructured difficulty distribution, making it infeasible to manually construct a task sequence in the order of difficulty.

We define a task instance as \textit{Success} if its episode length exceeds 900 alive steps and the EPTE-SP~$\tilde{\varepsilon}$ for both $\mathbf{v}_{b,x}^*$ and $\boldsymbol{\omega}_{b,z}^*$ remains below $30\%$.
We then identify the task instances classified as successful by LP-ACRL at 3000 iterations and compute the episodic rewards on this final success set.

\begin{figure}[h!]
    \centering
    \includegraphics[width=\linewidth]{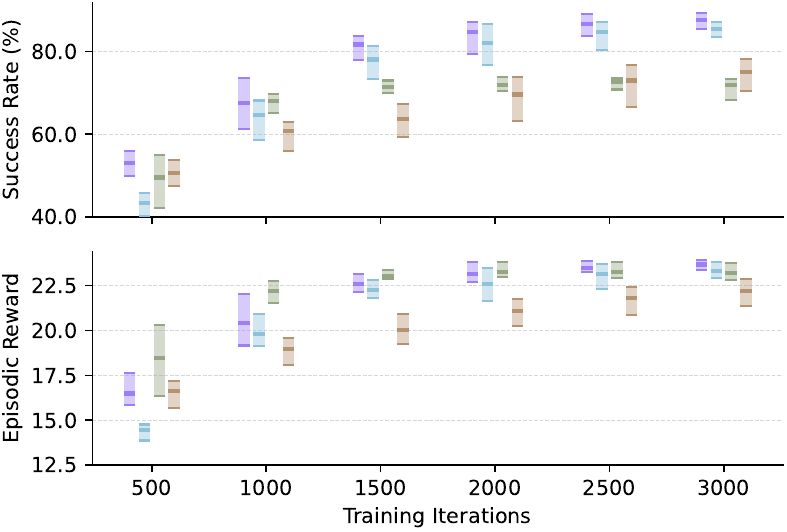}
    \includegraphics[width=\linewidth]{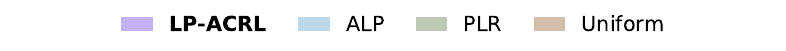}
    \caption{Evolution of Success Rate (top) and success set defined by LP-ACRL at 3000 training iterations. (bottom). \textbf{LP-ACRL} outperforms baselines in both metrics. The results confirm that our method satisfies the proposed objective: simultaneously minimizing the unmastered subspace via high success rates, and maximizing policy competence on mastered tasks via high rewards.}
    \label{fig:Successrate}
\end{figure}

Fig.~\ref{fig:Successrate} indicates that \textbf{LP-ACRL} reaches, within 1500 iterations, a success rate that most baselines do not achieve even after 3000 iterations, indicating superior sample efficiency.
\textbf{LP-ACRL} not only masters more tasks than all baselines, but also attains the highest average performance on the tasks it successfully learns. 
These results demonstrate that our \textbf{LP-ACRL} method performs effectively even under ill-defined difficulty structures, and scales well to large and complex task spaces.

We further deploy the student policy distilled from LP-ACRL on a real robot.
The height map is reconstructed using elevation mapping \cite{miki2022elevationmappinglocomotionnavigation}, which introduces noise on the real system, especially at higher velocities. To mitigate this effect, we adopt a teacher–student distillation framework, where the student policy integrates LSTM and MLP modules to exploit temporal information and improve robustness against noisy height maps.
The policy achieves linear velocities up to $3.0~\mathrm{m/s}$ on flat terrain and $2.5~\mathrm{m/s}$ on challenging terrains such as stairs, slopes, and gravel, as well as angular velocities up to $3.0~\mathrm{rad/s}$ across all terrains.

\section{Conclusion}
In this work, we present \emph{Learning Progress-based Automatic Curriculum Reinforcement Learning} (LP-ACRL), a simple and scalable framework that scales robot learning by estimating learning progress online from episodic rewards and using it to adapt task-sampling distributions.
By removing the need for hand-designed curriculum from difficulty ordering or manually tuned progression rules, LP-ACRL enables automatic curriculum generation in wide-ranging, multi-axis task spaces where difficulty is intractable or non-intuitive.

Across three locomotion task spaces---multi-level velocity tracking on flat terrain, multi-type rough terrain traversal, and a large-scale heterogeneous setting---LP-ACRL consistently outperforms handcrafted and automatic curriculum baselines in both convergence speed and final competence. In particular, in the scaled locomotion task space of 600 task instances, LP-ACRL masters a larger portion of the task set and achieves higher performance than existing methods on the solvable tasks, demonstrating strong scalability.

Finally, policies trained with LP-ACRL transfer successfully to the physical ANYmal D platform, achieving unprecedented high-speed and robust locomotion (up to $3.0~\mathrm{m/s}$ on flat terrain and $2.5~\mathrm{m/s}$ on challenging terrains, with up to $3.0~\mathrm{rad/s}$ angular velocity). These results support the broader view that learning-progress-driven task selection is an effective principle for scaling curriculum reinforcement learning from structured benchmarks to complex real-world robot learning task spaces.


\section*{ACKNOWLEDGMENT}
This research was supported by the ETH AI Center and the Swiss National Science Foundation through the National Centre of Competence in Automation (NCCR automation).

\appendix

\subsection{Locomotion Training Framework}
\label{apd:locomotion framework}
The settings largely follow the rough terrain locomotion training framework introduced by Rudin et al.~\cite{rudinLearningWalkMinutes2022} and Schwarke et al.~\cite{schwarke2025rslrl}.
\subsubsection{Action Space}
The action space is 12-dimensional, corresponding to the 12 joints of the robot. Each action represents the desired position of a joint.

\subsubsection{Observation Space}
The observation space includes the base linear and angular velocities, gravity vector measurements, joint positions and velocities, previous policy actions, and 108 terrain height samples collected from a grid centered at the robot base. The grid spans \(2.4\,\mathrm{m}\) along the \(x\)-axis and \(1.0\,\mathrm{m}\) along the \(y\)-axis, with a resolution of \(0.1\,\mathrm{m}\).  
Table~\ref{tab:observations} lists the observation dimensions.
\begin{table}[h]
\centering
\caption{Observation space dimensions for the locomotion task.}
\label{tab:observations}
\begin{tabular}{ll}
\toprule
\textbf{Observation} & \textbf{Dimension} \\
\midrule
Base linear velocity & 3 \\
Base angular velocity & 3 \\
Gravity vector & 3 \\
Velocity commands & 3 \\
Joint position & 12 \\
Joint velocity & 12 \\
Previous actions & 12 \\
Height map & 275 \\
\bottomrule
\end{tabular}
\end{table}
In the Multi-Level Linear-Velocity Tracking on Flat Terrain Experiment, the height map is excluded from the observations.
\subsubsection{Reward Functions}
Table~\ref{tab:rewardssymbols} summarizes the variables used in the reward functions, and Table~\ref{tab:rewardfunctions} lists the definitions and corresponding weights of all reward terms.

\begin{table}[h]
\centering
\caption{Definition of symbols in the reward functions.}
\label{tab:rewardssymbols}
\begin{tabular}{rl}
\toprule
Joint positions & $\mathbf{q}_j$ \\
Joint velocities & $\dot{\mathbf{q}}_j$ \\
Joint accelerations & $\ddot{\mathbf{q}}_j$ \\
Target joint positions & $\ddot{\mathbf{q}}_j^*$ \\
Joint torques & $\boldsymbol{\tau}_j$ \\
Base linear velocity & $\mathbf{v}_b$ \\
Base angular velocity & $\boldsymbol{\omega}_b$ \\
Commanded base linear velocity & $\mathbf{v}_b^*$ \\
Commanded base angular velocity & $\boldsymbol{\omega}_b^*$ \\
Number of collisions & $n_c$ \\
Feet air time & $\mathbf{t}_{air}$ \\
Environment time step & $dt$ \\
\bottomrule
\end{tabular}
\end{table}

\begin{table}[h]
\centering
\caption{Reward function definitions and weights}
\label{tab:rewardfunctions}
\begin{tabular}{lll}
\toprule
\textbf{Term} & \textbf{Definition} & \textbf{Weight} \\
\midrule
Linear velocity tracking & $\phi(\mathbf{v}_{b,xy}^*-\mathbf{v}_{b,xy})$ & $1\,dt$ \\
Angular velocity tracking & $\phi(\boldsymbol{\omega}_{b,z}^*-\boldsymbol{\omega}_{b,z})$ & $0.5\,dt$ \\
Vertical velocity penalty & $-\mathbf{v}_{b,z}^2$ & $4\,dt$ \\
Angular velocity penalty & $-\|\boldsymbol{\omega}_{b,xy}\|^2$ & $0.05\,dt$ \\
Joint motion penalty & $-\|\ddot{\mathbf{q}}_j\|^2-\|\dot{\mathbf{q}}_j\|^2$ & $0.001\,dt$ \\
Joint torque penalty & $-\|\boldsymbol{\tau}_j\|^2$ & $0.00002\,dt$ \\
Action rate penalty & $-\|\mathbf{q}_j^*\|^2$ & $0.25\,dt$ \\
Collision penalty & $-n_{collision}$ & $0.001\,dt$ \\
Feet air time reward & $\sum_{f=0}^4(\mathbf{t}_{air,f}-0.5)$ & $2\,dt$ \\
\bottomrule
\end{tabular}
\end{table}
Here, $\phi(x) = \exp(-\|x\|^2 / 0.25)$. The $z$-axis is aligned with gravity. 
\subsection{Baseline Methods}
\label{apd:baseline methods}
\subsubsection{Absolute Learning Progress}
Absolute Learning Progress(ALP)~\cite{portelasTeacherAlgorithmsCurriculum2019} is computed as:
\begin{align}
ALP_{c_j}(\zeta)=\lvert R_{c_j}(\zeta)-R_{c_{j-1}}(\zeta)\rvert.
\end{align}
And the task-sampling distribution is given by:
\begin{align}
  c(\zeta)=\frac{e^{AL P(\zeta) / \beta}}{\sum_{\zeta^{\prime} \in \tilde{\mathcal{T}}} e^{AL P\left(\zeta^{\prime}\right) / \beta}}.  
\end{align}

\subsubsection{Prioritized Level Replay}
We adopt the Prioritized Level Replay (PLR) framework \cite{schulman2018highdimensionalcontinuouscontrolusing}, which uses the agent's prediction error as the task sampling criterion. The score $S_\zeta$ for each task instance is computed following the Generalized Advantage Estimate (GAE), and the sampling probability is given by
\begin{align}
c(\zeta)=\frac{e^{S_\zeta / \beta}}{\sum_{\zeta^{\prime} \in \tilde{\mathcal{T}}} e^{S_{\zeta^\prime} / \beta}} .
\end{align}

\subsubsection{Low-Reward Prioritized Curriculum}
The Low-Reward Prioritized Curriculum (LRPC) assigns higher sampling probabilities to tasks with lower episodic rewards:
\begin{align}
  c(\zeta)=\frac{e^{-R(\zeta) / \beta}}{\sum_{\zeta^{\prime} \in \tilde{\mathcal{T}}} e^{-R\left(\zeta^{\prime}\right) / \beta}}.  
\end{align}

\subsubsection{Simple Hand-Crafted Curriculum}
This hand-crafted curriculum~\cite{jiConcurrentTrainingControl2022} progressively enlarges the sampling range of the velocity command throughout training.
Specifically, the maximum forward velocity, $v_{x, \max}$, increases with the training iteration, $k$, following the schedule:
\begin{align}
\lvert v_{x,\max} \rvert
= 1 + \frac{3.0}{1 + \exp\big(-0.002\,(k - 1000)\big).}
\label{eq:vxmax}
\end{align}

\subsection{Terrain Parameters for Scaled Locomotion Experiment}
\label{apd:terrain parameters}
\begin{table}[h]
\centering
\label{tab:terrain_params}
\resizebox{\linewidth}{!}{%
\begin{tabular}{lccc}
\toprule
\textbf{Terrain Type} & \textbf{Parameter Set} & \textbf{Level 0 (Min)} & \textbf{Level 3 (Max)} \\
\midrule
Ascending Stairs & step height [m] & 0.05 & 0.20 \\
Descending Stairs & step height [m] & 0.05 & 0.20 \\
Uphill Slope & slope gradient [-] & 0.00 & 0.40 \\
Downhill Slope & slope gradient [-] & 0.00 & 0.40 \\
Random Rough Terrain & -- & -- & -- \\
\bottomrule
\end{tabular}%
}
\end{table}
For random rough terrain, each level corresponds to a height field generated by uniformly sampling elevations within $0.02-0.10~$m.


\bibliographystyle{ieeetr}
\bibliography{ref}

\end{document}